%% file: Knowledge_distillation_survey.tex
\documentclass[conference]{IEEEtran}
\usepackage[font=footnotesize]{caption}
\IEEEoverridecommandlockouts

\usepackage{natbib}
\usepackage{tikz}
\usepackage{float}
\usepackage{forest}
\usetikzlibrary{arrows.meta, shapes.geometric, positioning, fit, backgrounds}
\usepackage{booktabs}
\usepackage{subcaption}

\usepackage{amsmath,amssymb,amsfonts}
\usepackage{algorithmic}
\usepackage{graphicx}
\usepackage{textcomp}
\usepackage{xcolor}
\usepackage{pgfplots}
\pgfplotsset{compat=1.18}

\usepackage{hyperref}
\hypersetup{
    colorlinks=true,
    linkcolor=darkblue,
    citecolor=darkblue,
    urlcolor=darkblue
}
\definecolor{darkblue}{rgb}{0.0, 0.0, 0.55}

\def\BibTeX{{\rm B\kern-.05em{\sc i\kern-.025em b}\kern-.08em T\kern-.1667em\lower.7ex\hbox{E}\kern-.125emX}}

\begin{document}

\title{Feature Alignment and Representation Transfer in Knowledge Distillation for Large Language Models%
\thanks{\hspace*{-\parindent}\rule{3.8cm}{0.4pt} \\ 
$\ast$: Equal contribution. \\
$\dagger$: Corresponding author.}
}

\author{
    \IEEEauthorblockN{
        Junjie Yang$^{\ast}$\textsuperscript{1},
        Junhao Song$^{\ast\dagger}$\textsuperscript{2,12},
        Xudong Han$^{\ast}$\textsuperscript{3},
        Ziqian Bi$^{\ast}$\textsuperscript{4,12},
        Tianyang Wang\textsuperscript{5,12},
        Chia Xin Liang\textsuperscript{6,12},
    }
    \IEEEauthorblockN{
        Xinyuan Song\textsuperscript{7,12},
        Yichao Zhang\textsuperscript{8,12},
        Qian Niu\textsuperscript{9,12},
        Benji Peng\textsuperscript{10,12},
        Keyu Chen\textsuperscript{11,12},
        Ming Liu\textsuperscript{4,12}
    }
    \\
    \IEEEauthorblockA{\textsuperscript{1}Pingtan Research Institute of Xiamen University, youngboy@xmu.edu.cn}
    \IEEEauthorblockA{\textsuperscript{2}Imperial College London, junhao.song23@imperial.ac.uk}
    \IEEEauthorblockA{\textsuperscript{3}University of Sussex, xh218@sussex.ac.uk}
    \IEEEauthorblockA{\textsuperscript{4}Purdue University, bi32@purdue.edu, liu3183@purdue.edu}
    \IEEEauthorblockA{\textsuperscript{5}University of Liverpool, tianyangwang0305@gmail.com}
    \IEEEauthorblockA{\textsuperscript{6}JTB Technology Corp., marcus.chia@ai-agent-lab.com}
    \IEEEauthorblockA{\textsuperscript{7}Emory University, xsong30@emory.edu}
    \IEEEauthorblockA{\textsuperscript{8}The University of Texas at Dallas, yichao.zhang.us@gmail.com}
    \IEEEauthorblockA{\textsuperscript{9}Kyoto University, niu.qian.f44@kyoto-u.jp}
    \IEEEauthorblockA{\textsuperscript{10}AppCubic, benji@appcubic.com}
    \IEEEauthorblockA{\textsuperscript{11}Georgia Institute of Technology, kchen637@gatech.edu}
    \IEEEauthorblockA{\textsuperscript{12}AI Agent Lab, ai-agent-lab.com}
}

\maketitle

\begin{abstract}
Knowledge distillation (KD) is a technique for transferring knowledge from complex teacher models to simpler student models, significantly enhancing model efficiency and accuracy. It has demonstrated substantial advancements in various applications including image classification, object detection, language modeling, text classification, and sentiment analysis. Recent innovations in KD methods, such as attention-based approaches, block-wise logit distillation, and decoupling distillation, have notably improved student model performance. These techniques focus on stimulus complexity, attention mechanisms, and global information capture to optimize knowledge transfer. In addition, KD has proven effective in compressing large language models while preserving accuracy, reducing computational overhead, and improving inference speed. This survey synthesizes the latest literature, highlighting key findings, contributions, and future directions in knowledge distillation to provide insights for researchers and practitioners on its evolving role in artificial intelligence and machine learning.
\end{abstract}


\section{Introduction}

\input{kd}

Knowledge Distillation (KD) has emerged as a fundamental technique in deep learning, enabling the transfer of knowledge from a complex and pretrained model (the teacher model) to a simpler and smaller model (the student model) \citep{hinton2015distillingknowledgeneuralnetwork, 2304.04615}. This approach is effective for model compression, significantly improving their efficiency and making them more suitable for deployment on resource-limited devices. The core idea behind KD is to distill the dark knowledge embedded in the teacher model into a student model, which enhances student performance \citep{2411.01547}. Recent studies have focused on enhancing the effectiveness of KD by developing innovative variants, frameworks, and methodologies, including decoupling logit-based and feature-based knowledge distillation methods \citep{2410.14741}, correlation-aware knowledge distillation \citep{2109.12507}, and partial-to-whole knowledge distillation \citep{1912.13179}.

\begin{table}[t]
\centering
\caption{Qualitative Comparison of Teacher and Student Models.}
\label{tab:model_comparison}
\begin{tabular}{lccc}
\toprule
\textbf{Model Type} & \textbf{Size} & \textbf{Inference Speed} & \textbf{Memory Usage} \\
\midrule
Teacher Model & Large & Slow & High \\
Student Model & Small & Fast & Low \\
\bottomrule
\end{tabular}
\end{table}

A common theme among these studies is the emphasis on feature alignment, which is crucial for Knowledge Distillation (KD) \citep{2411.01547}. Aligning the features of teacher and student models enhances the knowledge transfer process, leading to improved model performance. As shown in Figure \ref{fig:distillation}, another important aspect of KD is the decomposition of knowledge into smaller components, which can facilitate more effective knowledge transfer \citep{2109.12507}. This approach has been explored in various studies, where methods for decomposing teacher knowledge into subnetworks with increasing channel width have been proposed. For example, recent studies have investigated novel strategies for splitting teacher models into smaller, more manageable components to improve student model efficiency \citep{2103.07350, 1912.10850}.

The development of a general framework or model for knowledge distillation (KD) remains an active area of research because it offers a unified perspective on diverse KD techniques and facilitates the systematic analysis and development of new strategies \citep{2302.00444}. Significant contributions have been made in this area, with innovative frameworks and methodologies proposed to enhance the efficiency and scalability of KD. Recent studies have explored the applicability of KD across various domains, including natural language processing \citep{2211.01071} and computer vision \citep{2006.01683}, demonstrating its versatility and broad potential for practical deployment. The complexity of various distillation methods can introduce overhead to the training process \citep{2410.14741}. In addition, the scalability and generalizability of KD techniques are critical considerations for practical applications \citep{2109.12507}. Furthermore, the choice of teacher-student architecture and the design of the distillation loss function can significantly influence KD performance \citep{1912.13179}. To address these challenges, further research is required to develop more efficient and robust KD techniques.

In summary, knowledge distillation has become a crucial component of deep learning research, offering a wide range of applications and potential benefits. By examining the key themes and developments in the field, researchers can build on the existing work and push the boundaries of KD. As the field continues to evolve, it is essential to address the challenges and limitations associated with KD, ultimately leading to more efficient, effective, and scalable knowledge transfer techniques \citep{2311.13811}. This paper presents a survey of recent advancements in KD, providing a comprehensive overview of the current state of the field and highlighting its significant contributions, trends, and future directions.

\section{Background and Foundations}

\begin{figure*}[!t]
    \centering
    \includegraphics[width=\linewidth]{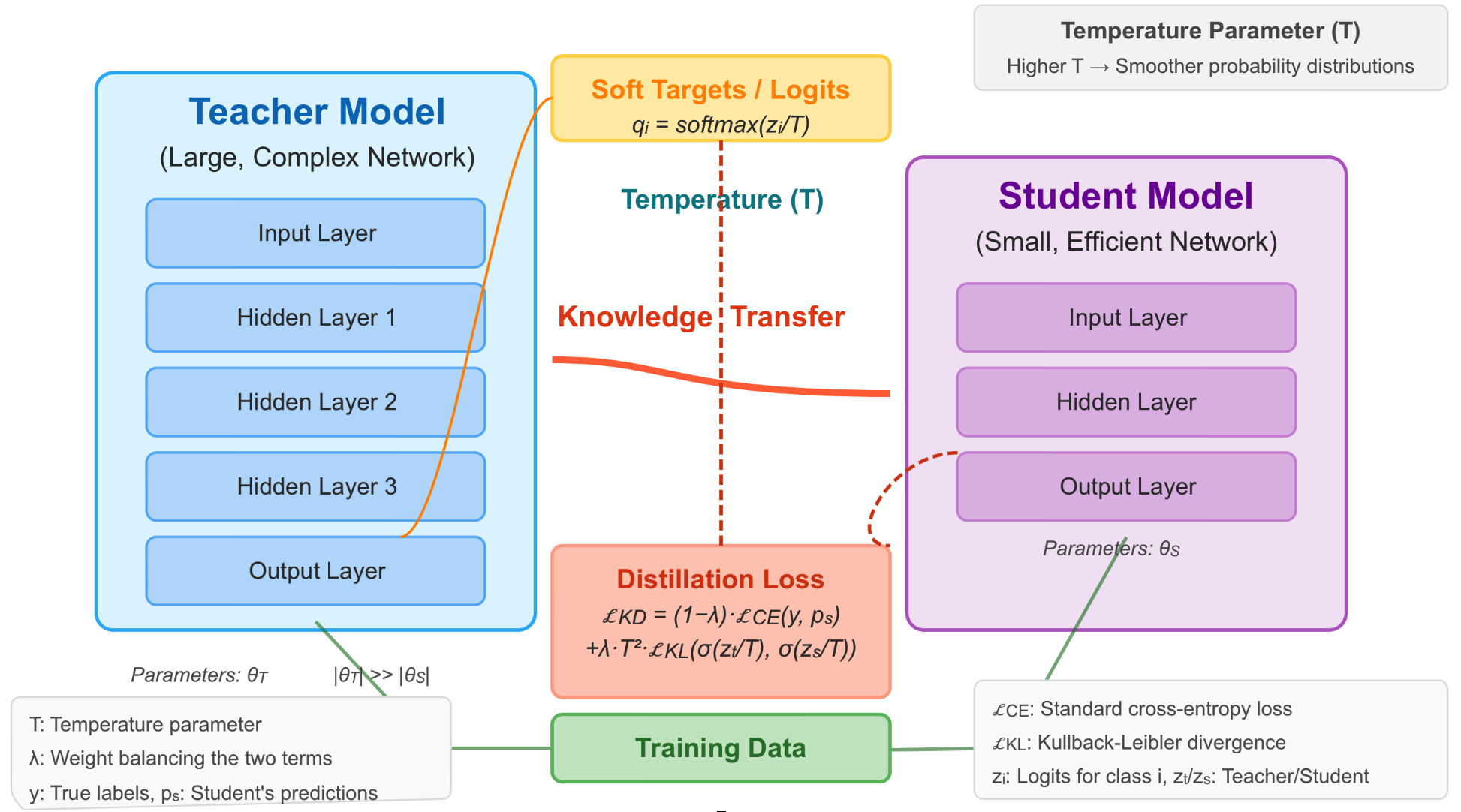}
    \caption{Foundations of Knowledge Distillation. The teacher model uses the temperature parameter \(T\) to produce softened predictions (soft targets) that capture nuanced information and guide the student model. The student is trained with a loss combining KL divergence between its outputs and the teacher’s soft targets, and cross-entropy with the true labels.}
    \label{found_of_kd}
\end{figure*}

The concept of knowledge distillation (KD) has gained significant attention in recent years, with numerous studies focusing on improving the effectiveness of this technique \citep{2011.14554, 2109.12507, 2304.04615}. The architecture of the knowledge distillation framework can be visualized, as shown in Figure~\ref{found_of_kd}, which illustrates the key components and their interactions in the distillation process. This framework typically involves a teacher model, which is pretrained on a large dataset, and a student model that learns from the teacher's output. This process allows the student to mimic the teacher's behavior and transfer knowledge in a more efficient manner, especially for applications where computational resources are limited.

Knowledge distillation (KD) involves transferring knowledge from a pre-trained teacher model to a student model, typically with a smaller capacity, to enhance student performance \citep{1912.10850}. This is typically achieved by optimizing a combined loss function that blends hard-target supervision with a softened probability distribution from the teacher model.
\begin{equation}
\mathcal{L}_{\text{KD}} = (1 - \lambda) \cdot \mathcal{L}_{\text{CE}}(\mathbf{y}, \mathbf{p}_s) + \lambda \cdot T^2 \cdot \mathcal{L}_{\text{KL}}\left(\sigma(\mathbf{z}_t/T), \sigma(\mathbf{z}_s/T)\right),
\end{equation}
where $\mathcal{L}_{\text{CE}}$ is the standard cross-entropy loss, $\mathcal{L}_{\text{KL}}$ is the Kullback-Leibler divergence \citep{hinton2015distillingknowledgeneuralnetwork}, $T$ is the temperature, $\lambda$ balances the two terms, and $\sigma(\cdot)$ denotes the softmax function.

This process has been shown to be highly effective in various applications including image classification, natural language processing, and speech recognition.
Recent studies have introduced new variants of knowledge distillation (KD), such as teaching-assistant distillation, curriculum distillation, mask distillation, and decoupling distillation, which aim to improve the performance of KD by introducing additional components or modifying the learning process \citep{2411.01547, 2410.14741}. Furthermore, novel frameworks, such as partial-to-whole knowledge distillation (PWKD) and Correlation-Aware Knowledge Distillation (CAKD) have been proposed to improve the efficacy of KD by decomposing knowledge and prioritizing influential distillation components \citep{2109.12507, 2410.14741}. These advancements have not only improved the accuracy of student models but have also provided new insights into the KD process.

A common theme observed across these studies is the emphasis on improving knowledge transfer, with a focus on decomposing knowledge to enhance the distillation process \citep{2109.12507, 2410.14741}. Feature alignment has also emerged as a crucial aspect of KD, with various approaches being explored, including logit-based and feature-based methods \citep{2112.10047, 2411.01547}. While some studies advocate logit-based methods, others focus on feature-based approaches, highlighting the complexity of selecting between these two paradigms \citep{2002.03532}. Exploration of different perspectives on feature alignment has led to a deeper understanding of the KD process and its applications.

The technical details and methodologies employed in these studies are diverse and innovative. For instance, PWKD decomposes knowledge by reconstructing the teacher into weight-sharing subnetworks by increasing channel width \citep{2109.12507}. By contrast, CAKD decouples the Kullback-Leibler divergence \citep{wu2024rethinking} into three unique elements: Binary Classification Divergence, Strong Correlation Divergence, and Weak Correlation Divergence \citep{2410.14741}. Block-wise logit distillation is another framework that applies an implicit logit-based feature alignment by gradually replacing teacher's blocks as intermediate stepping stone models \citep{2411.01547}. These methodologies have not only improved the accuracy of KD but have also provided new tools for analyzing and understanding the knowledge transfer process.

The performance gap between teacher and student models remains a major concern even with advanced KD techniques \citep{1912.10850}. Computational complexity is another issue, because some methods require significant resources owing to the decomposition of knowledge \citep{2109.12507}. Furthermore, choosing between logit-based and feature-based methods can be difficult because both have strengths and weaknesses \citep{2002.03532}. Addressing these challenges is crucial for advancing KD research and improving its applications.

The foundations of knowledge distillation (KD) have advanced significantly in recent years with a focus on improving knowledge transfer, decomposing knowledge, and feature alignment. Although challenges remain, developments in this field have far-reaching implications for various applications including image classification, natural language processing, and speech recognition \citep{2011.14554, 2304.04615}. As KD continues to evolve, it is essential to address the existing limitations and explore new methodologies to further enhance the effectiveness of this technique \citep{2410.14741, 2411.01547}. Ultimately, a deeper understanding of the KD process is crucial for unlocking its full potential and exploring new applications in various fields.

\section{Knowledge Distillation in Deep Learning}

Techniques for knowledge distillation (KD) in deep learning have been extensively explored in recent years, with a focus on transferring knowledge from complex models to smaller and more efficient models while preserving accuracy \citep{2007.09029, 2404.00936}. The teacher-student architecture is a common framework for KD, where the teacher model guides the student model to learn from its representations \citep{2011.14554, 2304.04615}. Various techniques have been proposed to improve KD performance, including teaching-assistant distillation, curriculum distillation, mask distillation, and decoupling distillation \citep{2007.09029}. In most approaches, soft labels are generated by applying temperature scaling to the teacher logits \citep{muller2019does}:

\begin{equation}
p_i = \frac{\exp(z_i / T)}{\sum_{j} \exp(z_j / T)},
\end{equation}
where $z_i$ denotes the teacher's logit for class $i$ and $T$ is the temperature used to soften the distribution.

These approaches aim to address the challenges of model compression and acceleration, which are crucial for deploying deep learning models on devices with limited resources \citep{2006.05525}.

The need for effective KD techniques has led to the development of new metrics for evaluating their performance, such as the distillation metric, which compares different algorithms based on their sizes and accuracy scores \citep{2007.09029}. Other evaluation metrics, including the accuracy, have also been widely used to assess the effectiveness of KD \citep{2004.08116, 2104.08448}. Recent studies have applied KD to various domains, including computer vision \citep{2404.00936} and natural language processing \citep{2310.02421}, thereby demonstrating its potential to improve the performance of smaller models.

Key findings and contributions in KD research include the proposal of new variants and techniques, such as dynamic rectification KD \citep{2201.11319}, data-efficient stagewise KD \citep{1911.06786}, and triplet loss for KD \citep{2004.08116}. These developments highlight the importance of transferring knowledge from larger to smaller models while preserving accuracy, which is a crucial aspect of KD \citep{2012.15495, 2306.10687}. Furthermore, research has focused on understanding what is lost in KD \citep{2311.04142} and quantifying knowledge in a deep neural network to explain KD for classification \citep{2208.08741}.

KD techniques have significant implications because they enable the deployment of accurate models on devices with limited resources, which is essential for real-world applications \citep{2006.05525}. However, challenges and limitations remain, including the need for further research to improve the effectiveness of KD and address the limitations of current techniques \citep{2011.14554, 2404.00936}. Despite these challenges, KD has emerged as a crucial technique in deep learning, with significant potential for improving model efficiency and accuracy \citep{2007.09029, 2304.04615}. As research continues to advance in this area, we expect to observe new developments and applications of KD techniques, leading to more efficient and accurate models in various domains \citep{2209.15555, 2403.05894, 2412.11788}.

\section{Knowledge Distillation in Computer Vision}

Applications of knowledge distillation (KD) in computer vision have been extensively explored in recent years, with a focus on improving the efficiency and performance of deep learning models. According to recent research, KD has been shown to significantly improve the performance of smaller and simpler models, making them more suitable for deployment in resource-constrained environments \citep{2404.00936}. This is particularly important in computer vision tasks such as image classification, object detection, and low-resolution image recognition, where model size and computational requirements are major concerns.

Recent variants of KD, such as teaching-assistant distillation, curriculum distillation, mask distillation, and decoupling distillation, have also shown promising results in improving KD performance \citep{2304.04615}. These methods have been applied to various computer vision tasks, including image classification \citep{2004.05937} and object detection \citep{2308.09105}, with impressive results. Furthermore, the use of a quantized embedding space for transferring knowledge has yielded state-of-the-art results in KD \citep{1912.01540}, highlighting the importance of efficient knowledge transfer mechanisms.

Another key development in KD is the concept of pixel distillation, which extends KD to the input level, allowing flexible cost control and improved performance in low-resolution image recognition \citep{2112.09532}. This approach has been shown to be effective in improving model performance for low-resolution images, making it a promising direction for future research. In addition, the use of frequency attention for KD has been explored, yielding promising results \citep{2403.05894}.

The choice of teacher and student architectures, as well as the size of the student model, can significantly affect the accuracy and efficiency of KD \citep{2407.12808}. This highlights the need for careful consideration of these factors when applying KD to computer vision tasks. Moreover, the evaluation metrics used to assess the performance of KD, such as the accuracy, inference speed, and computational workload, can also significantly impact the results \citep{2006.01683}.

Despite this progress, KD still has several limitations and challenges. For example, KD can be computationally expensive, particularly when using large teacher models or complex distillation methods \citep{2404.00936}. Additionally, KD can lead to performance degradation if not implemented carefully, such as by choosing an inappropriate teacher-student architecture or distillation method \citep{2304.04615}. Scalability remains a concern because KD may not scale well to very large models or complex tasks, necessitating further research to develop more efficient and effective methods \citep{1912.01540}.

In conclusion, the application of KD in computer vision has made significant progress, focusing on improving the efficiency and performance of deep learning models. The development of new techniques such as pixel distillation and frequency attention has shown promising results. However, careful consideration of teacher-student architectures and evaluation metrics is crucial for achieving optimal results. However, several limitations and challenges remain, highlighting the need for further research. KD has the potential to play a key role in the development of efficient and effective computer vision models, and ongoing research is likely to lead to exciting future developments \citep{2004.05937}.

\section{Knowledge Distillation for Natural Language Processing Tasks}

Knowledge distillation (KD) for natural language processing (NLP) tasks has garnered significant attention in recent years \citep{huang2024leveraging} with a plethora of research focused on improving the efficiency and accuracy of large language models \citep{yang2024survey}. The introduction of open-source toolkits such as TextBrewer \citep{2002.12620} has facilitated the setup of distillation experiments across various neural network models and NLP tasks, offering flexibility and adaptability in the KD process. Surveys on recent teacher-student learning studies \citep{2304.04615} and knowledge distillation for large language models \citep{2407.01885} have highlighted the diversity of methods being explored, including white-box \citep{nguyen2022black}, black-box knowledge distillation \citep{liang2021distill}, teaching-assistant distillation, curriculum distillation, mask distillation, and decoupling distillation.

A key theme emerging from these studies is the importance of balancing the efficiency and accuracy of the KD. Researchers have sought to improve the efficiency of large language models by reducing their size or improving inference speed without significantly compromising their accuracy \citep{2006.00844, 2106.04563}. The development of novel distillation methods, such as zero-shot knowledge distillation \citep{2012.15495}, has also addressed the challenges related to data accessibility and privacy, enabling student networks to learn from teachers without access to task-specific data.

The choice between sentence-level and token-level distillation methods has been a subject of debate, with studies suggesting that token-level distillation is more suitable for simple scenarios, whereas sentence-level distillation excels in complex scenarios \citep{2404.14827}. Hybrid approaches combining both sentence-level and token-level distillation have also been proposed, offering a potential solution to scenario-dependent preferences \citep{2404.14827}. Furthermore, the use of homotopic task-agnostic distillation \citep{2302.09632} and cohesive distillation architectures \citep{2301.08130} have been explored to improve the efficiency and effectiveness of KD in NLP tasks.

Despite these advancements, challenges persist, including the trade-off between efficiency and accuracy, data accessibility, and defining complexity levels for scenario-dependent method selection \citep{2109.05696}. A survey of recent teacher-student learning studies \citep{2304.04615} and an analysis of knowledge distillation for large language models \citep{2407.01885} have underscored the need for continued innovation to address these challenges. Future research directions may include exploring multi-stage balanced distillation \citep{2406.13114}, f-divergence minimization for sequence-level knowledge distillation \citep{2307.15190}, and multi-task learning approaches \citep{1910.10683} to further improve the efficiency and accuracy of large language models.

The implications of these developments are far-reaching, with potential applications in natural language processing, machine translation, question answering, and text classification. As KD continues to evolve, it is likely to play an increasingly important role in enabling the efficient deployment of large language models in real-world applications \citep{abdullah2022chatgpt}, while also addressing pressing challenges related to data privacy and accessibility. Ultimately, the connections between KD, efficiency, and accuracy remain a critical area of research, driving innovation and advancements in the field of natural language processing \citep{hahn2019self}.

\section{Efficient Training and Hyperparameter Tuning of Knowledge Distillation}

Efficient training runs and hyperparameter tuning are crucial aspects of knowledge distillation (KD) because they significantly affect the performance and computational efficiency of the student model. Recent studies have made notable contributions in this area by providing valuable insights into the development of efficient KD techniques. For instance,  \citep{2303.06480} demonstrated that augmenting future training runs with KD from previous runs can reduce the training time by 80-90\% with a minimal effect on accuracy. This approach highlights the potential for significant computational savings through the strategic application of the KD. The concept of weight distillation, as introduced by \citep{2009.09152}, has also shown promise for reducing training time while maintaining competitive performance. By transferring knowledge in the neural network parameters through a parameter generator, weight distillation reduces the training time by 1.88-2.94x. Furthermore, self-distillation methods, such as those proposed by \citep{2103.07350}, eliminate the need for pre-trained teacher models, thereby reducing the computational and storage overhead. These developments underscore the importance of exploring innovative KD techniques to improve the efficiency and reduce the computational costs.

Hyperparameter tuning is another critical aspect of KD, as has been emphasized by  \citep{2309.03659}. This study highlights the significance of hyperparameter tuning for achieving competitive results by establishing a solid baseline for comparing KD techniques in semantic image segmentation. The importance of sufficient hyperparameter optimization was further reinforced by Ruffy and Chahal \citep{1912.10850}, who demonstrated that inadequate tuning can lead to underperformance of student models. Therefore, it is essential to devote considerable attention to hyperparameter tuning when implementing the KD techniques.

The papers analyzed also revealed common themes and patterns, including the utilization of KD as a means of transferring knowledge from a large model to a smaller one, improving efficiency, and reducing computational costs. Efficient training methods such as sequence-based training, weight distillation, and self-distillation have been reported in the literature. Additionally, hyperparameter tuning has been consistently identified as a crucial aspect of KD, with some studies providing extensive information on the optimization procedures.

However, contrasting viewpoints and approaches are evident in literature. For example, traditional teacher-student learning, as discussed in  \citep{2304.04615}, is contrasted with novel frameworks that eliminate the need for pretrained teacher models, such as self-distillation \citep{2103.07350}. The exploration of different distillation methods, including weight distillation, self-distillation, and various loss terms in semantic image segmentation, further highlights the diversity of the approaches in this area. The technical details and methodologies employed in these studies include the proposal or utilization of various KD algorithms such as weight distillation, self-distillation, and teacher-student learning \citep{2304.04615, 2112.10047, 2410.14741}. Experimental evaluations of diverse tasks and benchmark datasets are also commonly used to demonstrate the effectiveness of the proposed methods. Moreover, some papers emphasize the importance of hyperparameter tuning, providing detailed information on the optimization procedures. The computational overhead introduced by KD can be mitigated through novel frameworks or optimization techniques, as demonstrated in \citep{2307.08436}. Insufficient hyperparameter tuning can lead to the underperformance of student models, highlighting the need for thorough optimization procedures. Furthermore, the lack of standardized baselines and evaluation protocols hinders the comparability of results across different KD techniques, making it essential to establish common benchmarks and evaluation metrics.

\section{Semiparametric Inference and Foundations}

\begin{figure*}[!t]
    \centering
    \includegraphics[width=\linewidth]{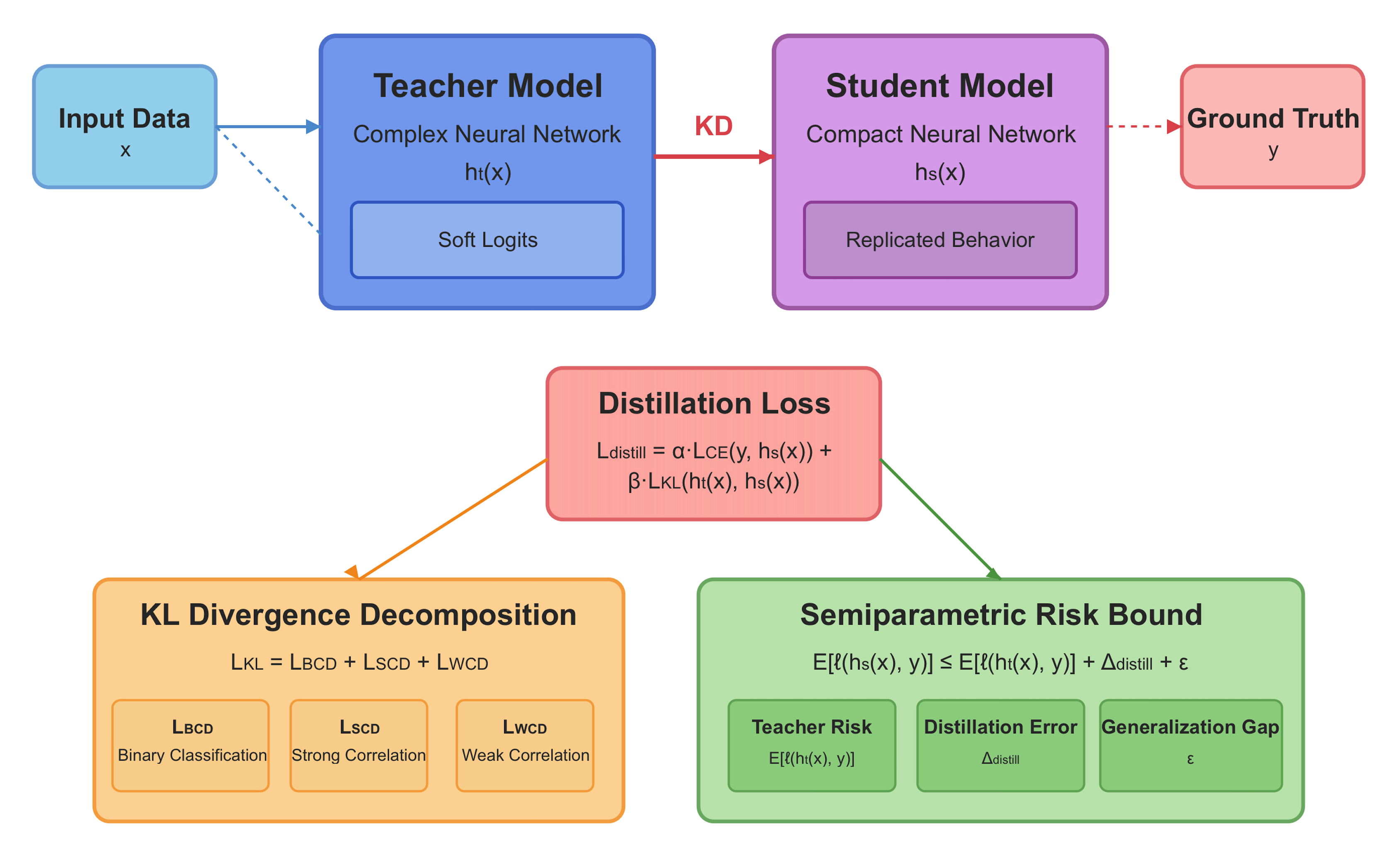}
    \caption{Semiparametric framework for Knowledge Distillation. The diagram illustrates how a complex teacher model transfers knowledge to a compact student model through distillation loss, with theoretical guarantees provided by KL-divergence decomposition and semiparametric risk bounds.}
    \label{Semiparametric_Framework}
\end{figure*}

The theoretical foundations of knowledge distillation (KD) have been extensively explored in recent years, with a growing body of research focusing on semiparametric inference as a framework for understanding this complex phenomenon. \citep{2104.09732} noted that casting KD as a semiparametric inference problem provides new guarantees for the prediction error of standard distillation and leads to enhancements, such as cross-fitting and loss correction. This perspective has been further reinforced by surveys, such as \citep{2011.14554}, which highlight the versatility and applications of KD beyond model compression, including learning using privileged information and generalized distillation. Figure~\ref{Semiparametric_Framework} presents a comprehensive semiparametric framework for KD, illustrating the key components of the teacher-student model architecture, loss function decomposition, and theoretical risk bounds that provide rigorous guarantees for distillation performance.

The state-of-the-art in KD for classification was thoroughly examined by \citep{1912.10850}, with experiments revealing that many state-of-the-art distillation results are difficult to reproduce, and classical distillation with data augmentation training schemes can provide orthogonal improvements. By contrast, \citep{2209.15555} presented a unified view of affinity-based KD, providing a framework for understanding relation-based algorithms and offering insights into effective design choices. Furthermore, \citep{2410.14741} introduced a correlation-aware KD framework based on decoupling the Kullback-Leibler divergence into unique elements and prioritizing the influential facets of the distillation components. Specifically, the Kullback-Leibler divergence between the teacher and student can be decoupled as:
\begin{equation}
\mathcal{L}_{\text{KL}} = \mathcal{L}_{\text{BCD}} + \mathcal{L}_{\text{SCD}} + \mathcal{L}_{\text{WCD}},
\end{equation}
where $\mathcal{L}_{\text{BCD}}$, $\mathcal{L}_{\text{SCD}}$, and $\mathcal{L}_{\text{WCD}}$ represent binary classification divergence, strong correlation divergence, and weak correlation divergence, respectively.

A common theme in these studies is the importance of understanding distillation components to optimize knowledge transfer. As emphasized by \citep{2309.03659}, KD is viewed as a flexible framework that can be applied to various problems beyond model compression. The role of semiparametric inference in providing a theoretical foundation for understanding KD cannot be overstated, thus leading to new guarantees and enhancements \citep{2104.09732}. From a semiparametric perspective, the expected risk of a student can be bounded as:
\begin{equation}
\mathbb{E}_{\mathcal{D}}[\ell(h_s(x), y)] \leq \mathbb{E}_{\mathcal{D}}[\ell(h_t(x), y)] + \Delta_{\text{distill}} + \varepsilon,
\end{equation}
where $\Delta_{\text{distill}}$ captures the error introduced by distillation and $\varepsilon$ is a generalization gap term.

Moreover, the technical details and methodologies used in these studies, such as modern semiparametric tools and decoupling Kullback-Leibler divergence, have enabled the development of novel frameworks and exploration of new applications. However, despite significant progress in this area, several challenges and limitations remain. As highlighted by Ruffy and Chahal \citep{1912.10850}, it is difficult to reproduce state-of-the-art distillation results, highlighting the need for more rigorous evaluation. In addition, some distillation techniques may only succeed in specific architectures and training settings, thereby limiting their applicability \citep{2203.10163}. The need for further research on KD is evident, including the development of new frameworks and the exploration of novel applications \citep{2307.11030}. Ultimately, a deeper understanding of the theoretical foundations of KD is crucial to unlocking its full potential and addressing the challenges that lie ahead.

The connection between KD and other research areas was noteworthy. For example, \citep{2102.11638} explored the relationship between KD and adversarial training, and \citep{2310.02421} examined the performance of student models in relation to their teachers. Furthermore, \citep{2305.17581} discussed the role of KD in partial variance reduction, highlighting its potential applications in areas such as uncertainty estimation. As the field continues to evolve, new and innovative connections are likely to be discovered, further solidifying the importance of KD in the broader landscape of machine learning research.

\section{Label Assistance and Beyond}

Recent years have witnessed significant advancements in knowledge distillation (KD) and novel approaches have been proposed to improve its efficiency and effectiveness. One such approach is Label Assisted Distillation (LAD), introduced by \citep{2407.13254}, which leverages label assistance to enhance the performance of a teacher model without requiring complex models or extra sensors. This method has shown promising results, particularly in semantic segmentation tasks. By contrast, \citep{2303.08360} investigated the effectiveness of classical KD techniques for multilabel classification and found that logit-based methods are not well suited for this task. Instead, the authors propose a novel distillation method based on Class Activation Maps (CAMs), which is effective and straightforward to implement for multi-label classification.

The use of unlabeled mismatched images for KD was also explored in \citep{1703.07131}, demonstrating the effectiveness of this approach in image classification networks. This line of research highlights the potential of KD to leverage diverse data sources to improve the model performance. Furthermore, \citep{2106.01489} and \citep{2302.08771} proposed explicit and implicit KD methods using unlabeled data, demonstrating the versatility of KD techniques. The compatibility between label smoothing and KD was revisited by \citep{2206.14532}, providing new insights into the relationship between these two techniques.

A notable theme emerging from these studies is the focus on improving teacher model performance either through label assistance or by employing more effective distillation methods. For instance, \citep{2407.13254} and \citep{2303.08360} aimed to enhance the teacher model performance, albeit using different approaches. Another common thread is the exploration of new distillation methods, such as CAM-based distillation and knowledge distillation, using unlabeled mismatched images. These developments underscore ongoing efforts to expand the scope and applicability of KD.

The technical details and methodologies employed in these studies vary, with \citep{2407.13254} utilizing a label noising approach to boost teacher performance, and \citep{2303.08360} leveraged CAMs to convey compact information about multiple labels simultaneously. The use of unlabeled mismatched images by \citep{1703.07131} highlights the importance of stimulus complexity in achieving good performance. These diverse methodologies demonstrate the richness and complexity of KD research.

However, classical KD techniques have been found to be less effective for multi-label classification \citep{2303.08360}. In addition, some KD approaches require complex teacher models or additional sensors, which can be a limitation \citep{2407.13254}. The lack of generalization in KD methods is another challenge because different methods are effective for different tasks and datasets \citep{2106.01489}. These limitations underscore the need for continued research on robust and adaptable KD techniques.

The implications of these developments are significant as they highlight the potential of KD to improve model performance in a variety of tasks, from image classification to semantic segmentation. The connections between KD and other techniques such as label smoothing are also being explored, providing new insights into the relationships between these methods \citep{2206.14532}. As research in this area continues to evolve, it is likely that we will see more innovative applications of KD, driving progress in the field of machine learning and beyond.

\input{timeline}

\section{Class Activation Maps and Feature-based}

Class Activation Maps (CAMs) and feature-based distillation methods have garnered significant attention in recent years, particularly in the context of knowledge distillation (KD) for multi-label classification scenarios. \citep{2303.08360} propose a novel distillation method based on CAMs, which has been shown to be both effective and straightforward to implement, outperforming other methods in multi-label classification settings. This approach highlights the potential of using CAMs to improve KD, particularly when dealing with complex scenarios where traditional logit-based methods may not perform well.

Another significant development in this area is the introduction of block-wise logit distillation \citep{2411.01547}. This framework applies an implicit logit-based feature alignment, achieving comparable or superior results to those of state-of-the-art distillation methods. The use of block-wise logit distillation demonstrates the effectiveness of hybrid approaches that combine logit-based and feature-based distillation methods, which is a common theme in recent studies.

A more recent study extends logit distillation to both the instance-level and class-level, enabling student models to capture higher-level semantic information from teacher models. Referred to as Class-aware Logit Knowledge Distillation (CLKD) \citep{2211.14773}, this approach underscores the importance of class-level information in KD, particularly for multi-label classification tasks.

Recent studies have focused on enhancing the efficiency and effectiveness of KD. For instance,  \citep{2303.15678} introduced a training-free architecture search method to identify the optimal student architectures for a given teacher, attaining state-of-the-art performance with a substantial reduction in training time. This perspective contrasts with conventional training-based approaches, which have important implications for the computational efficiency and scalability of the KD methods.

A common theme in these studies emphasized feature alignment as a critical component of effective KD. Whether through CAMs, block-wise logit distillation, or class-aware logit knowledge distillation, the goal of improving the feature-level alignment between teacher and student models is paramount. This suggests that future research should continue to explore innovative methods for enhancing the feature alignment, potentially leading to more robust and efficient KD techniques.

The technical details of these methodologies vary, with approaches including gradient-based CAMs generation \citep{2303.08360}, hierarchical architectures for block-wise logit distillation \citep{2411.01547}, and novel class correlation losses to force student models to learn inherent class-level correlations \citep{2211.14773}. These diverse methodologies underscore the complexity and richness of the field, with different approaches being suited to various scenarios and applications.

Scalability is a significant concern because some methods may not be readily applicable to larger models or datasets. Computational efficiency is another issue, particularly for training-free approaches that require substantial resources for an architecture search. Furthermore, multi-label classification scenarios continue to pose difficulties for KD, highlighting the need for more effective and specialized methods in this area.

Overall, recent research on CAMs and feature-based distillation methods has made significant strides in advancing our understanding of KD, particularly for complex scenarios such as multi-label classification. The emphasis on hybrid approaches, feature alignment, and efficient methodologies highlights the key themes and developments in the field. As research continues to address the challenges and limitations of the current methods, we expect to see even more innovative and effective techniques that can further enhance the capabilities of KD in deep learning applications.

\section{Multi-Label Classification and Semantic Segmentation}

Knowledge distillation (KD) for multi-label classification and semantic segmentation has been an active area of research, and various approaches have been proposed to improve the performance of these tasks. A key finding in this domain is the effectiveness of label-assisted distillation (LAD), which enhances the performance of lightweight teacher models by incorporating noisy labels into the input \citep{2407.13254}. This approach highlights the importance of utilizing auxiliary information, such as labels, to enhance the distillation process, and class activation map (CAM)-based distillation has been shown to be an effective method for multi-label classification, conveying class-wise information that is highly correlated with the final classification results \citep{2303.08360}.

Another significant development, multi-label knowledge distillation edge distillation, exploits informative semantic knowledge from logits and enhances the distinctiveness of learned feature representations by leveraging the structural information of label-wise embeddings \citep{2308.06453}. This approach underscores the importance of learning effective feature representations for successful KD. Feature Augmented Knowledge Distillation (FAKD) is another method that explores data augmentation in the feature space for efficient distillation in semantic segmentation, resulting in an improved performance without significant overhead \citep{2208.14143}.

The use of auxiliary information and importance of feature representations are common themes in these studies. For instance, \citep{2407.13254,2308.06453,2208.14143} utilized additional information such as labels, class activation maps, or feature augmentations to enhance the distillation process. Furthermore, \citep{2308.06453,2208.14143,2303.08360} highlighted the significance of learning effective feature representations for a successful KD. However, contrasting viewpoints and approaches exist in this domain. For example, \citep{2303.08360} found that logit-based methods are not well suited for multi-label classification, whereas \citep{2308.06453} proposed a method that exploits logits and feature representations. Additionally, \citep{2407.13254,2208.14143} used single teachers, whereas \citep{2208.10169} explored the use of multiple teachers in the form of feature augmentations.

From a technical perspective, various methodologies have been proposed to improve KD for multi-label classification and semantic segmentation. For instance, \citep{2407.13254} proposed a method for boosting the lightweight teacher performance by introducing label noise and using a dual-path consistency training strategy. \citep{2303.08360} uses CAMs to convey class-wise information for multi-label classification.  \citep{2208.14143} develops an approach to optimize feature augmentations in semantic segmentation, while \citep{2208.10169} proposes a multi-granularity distillation scheme with hierarchical loss setting and structurally complementary teachers.

\citep{2407.13254,2208.10169,2208.14143} highlight the need for efficient and scalable KD methods. \citep{2208.10169} addresses the challenge of limited labeled data in semi-supervised semantic segmentation, while \citep{2407.13254} and \citep{2308.06453} emphasize the importance of balancing teacher and student performance in KD.

In conclusion, KD for multi-label classification and semantic segmentation is a rapidly evolving field, and various approaches have been proposed to improve the performance of these tasks. The use of auxiliary information, such as labels and feature augmentation, and the importance of learning effective feature representations are the key themes in this domain. However, there are still limitations and challenges that need to be addressed, including scalability, efficiency, and the need for balanced teacher-student performance. Further research is needed to fully explore the potential of KD for these tasks and to address the remaining challenges in this domain. In addition, the connections between these approaches and other areas of research, such as transfer learning and few-shot learning, are worthy of further exploration \citep{1910.02771,2004.10934}.

\section{Open Challenges and Future Directions}

The field of knowledge distillation (KD) has witnessed significant advancements in recent years, with various studies exploring new variants, frameworks, and applications of this technique \citep{2304.04615,2410.14741}. Despite these developments, several open challenges and future directions remain to be addressed in KD research. One key finding that emerges from the analysis of relevant papers is the importance of improving KD performance, and many studies have focused on developing new methods to enhance the accuracy and efficiency of this process \citep{1912.10850}, \citep{2109.12507}. For instance, the introduction of correlation-aware KD frameworks has been shown to prioritize the most influential facets of distillation components, thereby leading to improved performance \citep{2410.14741}.

Another significant theme that emerges from the literature is the need for reproducibility and generalizability in KD research \citep{2011.14554, 1912.10850}. The lack of standardized evaluation protocols and benchmark datasets has hindered the comparability of results across different studies, highlighting the importance of establishing common evaluation frameworks to facilitate the development of more effective KD techniques \citep{2407.01885}. Furthermore, the versatility of KD has been demonstrated through its application to various frameworks and paradigms, including large language models \citep{2310.02421} and object detection \citep{2103.14337}.

The analysis of relevant papers also revealed contrasting viewpoints and approaches to KD, with some studies questioning the effectiveness of feature distillation methods and suggesting that classical distillation approaches may be more reliable \citep{1912.10850}. Additionally, decoupling distillation loss from task loss has been proposed as a means of improving knowledge transfer, and the CAKD framework provides a notable example of this approach \citep{2410.14741}. The use of novel architectures and frameworks such as teaching-assistant distillation and partial-to-whole knowledge distillation has also been explored in recent studies \citep{2304.04615,2109.12507}. The technical details and methodologies employed in KD research vary widely, with different loss functions, optimization techniques, and evaluation protocols used to facilitate knowledge transfer \citep{2112.10047,2008.01458}. The choice of architecture and training settings has also been shown to significantly affect the effectiveness of KD, highlighting the need for further research on the influence of these factors on the knowledge transfer process \citep{2205.02399}. Moreover, the development of new evaluation metrics and protection for assessment are crucial for assessing the performance of KD techniques in a more comprehensive manner \citep{2407.01885}. The lack of generalizability of many KD techniques to different architectures and training settings is a significant concern, which highlights the need for more robust and adaptable methods \citep{1912.10850}. Furthermore, standardized evaluation protocols and benchmark datasets are essential to facilitate the development of more effective KD techniques \citep{2407.01885}. Understanding the influence of distillation components on the knowledge transfer process is also crucial, and further research is required to elucidate the impact of these factors on KD performance \citep{2410.14741}.

KD is characterized by a range of open challenges and future directions, including the need for improved performance, reproducibility, and generalizability. The development of new variants, frameworks, and applications of KD has significant implications in various fields, including computer vision and natural language processing \citep{2304.04615,2310.02421}. Addressing the limitations and challenges identified in this survey is essential for advancing the field of KD and realizing its full potential in real-world applications \citep{2407.01885}. As research in this area continues to evolve, it is likely that new breakthroughs and innovations will emerge, further enhancing our understanding of KD and its role in facilitating effective knowledge transfer \citep{2410.14741}.

\section{Emerging Trends}

Knowledge distillation has undergone significant advancements in recent years, and various techniques and methodologies have been proposed to improve the efficiency and effectiveness of knowledge transfer from teacher to student models. As is evident from the analyzed papers, including \citep{2011.14554,2109.12507,2311.13811,2410.14741}, a common theme that emerges is the focus on improving knowledge transfer. Researchers have introduced new variants of KD such as teaching-assistant distillation, curriculum distillation, mask distillation, and decoupling distillation. These approaches aim to optimize knowledge transfer by introducing additional components or changing the learning process.

The development of correlation-aware KD frameworks, such as CAKD \citep{2410.14741}, has prioritized the facets of distillation components that have the most substantial influence on predictions, thereby optimizing the knowledge transfer from teacher to student models. Similarly, partial-to-whole knowledge distillation paradigms, such as PWKD \citep{2109.12507}, reconstruct the teacher into weight-sharing sub-networks with increasing channel width and train sub-networks jointly to obtain decomposed knowledge. Education distillation \citep{2311.13811} has also been proposed, which takes fragmented student models divided from the complete student model as lower-grade models and gradually integrates them into a complete target student model through dynamic incremental learning.

The analysis of these studies reveals that improving knowledge transfer, introducing new variants of KD, decomposing knowledge, and employing dynamic learning approaches are common themes and patterns in the field. However, contrasting viewpoints or approaches also exist, such as the focus on knowledge quality versus quantity \citep{2109.12507} and static versus dynamic learning \citep{2311.13811}. The technical details and methodologies employed in these studies include various neural network architectures, loss functions, and optimization algorithms, highlighting the diversity of approaches being explored in KD.

In addition, some limitations and challenges need to be addressed, including scalability, interpretability, and self-supervised learning. As highlighted in \citep{2011.14554}, KD is computationally expensive and may not be scalable to large-scale datasets. Furthermore, the need for explainable KD is emphasized, which can provide insights into the performance gain of KD \citep{1912.10850}. The potential of self-supervised learning in KD has also been mentioned, but this area remains underexplored \citep{2305.10893}.

The implications of these developments are significant because they have the potential to improve the performance and efficiency of deep neural networks in various applications including computer vision \citep{2410.12259}, natural language processing \citep{2310.02421}, and object detection \citep{2410.12259}. The connections between KD and other areas of research such as transfer learning and meta-learning have also been explored \citep{2112.10047}. As the field continues to evolve, it is likely that we will see further innovations and advancements in KD, leading to improved performance and efficiency of deep neural networks. Overall, the analysis of these papers highlights the importance of continued research on KD, with a focus on addressing the limitations and challenges, exploring new approaches, and investigating the connections to other areas of research.

\section{Reproducibility and Generalizability}

The concept of knowledge distillation (KD) has been extensively explored in recent years and various techniques have been proposed to improve its effectiveness. Detailed sections can be further divided into subsections, such as those focusing on the development of novel frameworks, such as CAKD \citep{2410.14741}, which prioritize the facets of distillation components with the most substantial influence on predictions, leading to optimized knowledge transfer. Another area of interest is the examination of the reproducibility and generalizability of KD techniques, as highlighted by the state of Knowledge Distillation for classification \citep{1912.10850}, suggesting that classical distillation combined with data augmentation can lead to orthogonal improvements.

Furthermore, research has also explored the use of intermediate models or teaching assistants to facilitate knowledge transfer, such as in modeling teacher-student techniques in deep neural networks \citep{1912.13179}, which provides a general model for KD, summarizing various methods and techniques, and facilitating the development of new strategies. In addition, dynamic incremental learning approaches such as Education Distillation \citep{2311.13811} have been proposed, where student models learn in a school-like setting, gradually improving performance. These developments highlight the importance of thoroughly examining each distillation component to optimize knowledge transfer and the need for reproducibility and generalizability of the KD techniques.

Other notable advancements include the application of KD to various domains, such as image classification \citep{2109.12507} and model compression \citep{2205.10003}, and the use of decoding techniques such as Decoupling Dark Knowledge via blockwise logit distillation for feature-level alignment \citep{2411.01547}. The technical details and methodologies employed in these approaches vary, with some studies, such as CAKD \citep{2410.14741}, decoupling the Kullback-Leibler divergence into unique elements, whereas others, such as The State of Knowledge Distillation for Classification \citep{1912.10850}, utilize standardized model architectures and fixed compute budgets to evaluate KD techniques.

The analysis of these papers also reveals contrasting viewpoints or approaches, with some focusing on introducing new techniques, such as CAKD \citep{2410.14741} and Education Distillation \citep{2311.13811}, while others, such as The State of Knowledge Distillation for Classification \citep{1912.10850}, emphasize the importance of evaluating and reproducing existing methods. Moreover, some approaches prioritize the use of complex models, such as teaching assistants, whereas others, such as classical distillation, rely on simpler architectures. These differences in methodology and approaches highlight the need for careful evaluation and comparison of different techniques to determine their effectiveness.

Moreover, KD still faces limitations and challenges, including the lack of generalizability and reproducibility of some techniques \citep{1912.10850}, complexity of introducing new components or modifying the learning process \citep{2311.13811}, and challenge of developing a unified framework that summarizes various KD methods and facilitates the development of new strategies \citep{1912.13179}. Nevertheless, the progress made in this field has significant implications for the development of more efficient and effective machine learning models, and ongoing research continues to explore new techniques and applications, such as Spot-adaptive Knowledge Distillation \citep{2205.02399}, Student-friendly Knowledge Distillation \citep{2305.10893}, and the Wisdom of the Crowds: Ensemble and Transfer Learning for Knowledge Distillation \citep{2401.01234}.

\section{Logit-based Knowledge Distillation}

\begin{figure*}[ht]
\centering
\begin{tikzpicture}[
  node distance=1.5cm and 1.8cm,
  box/.style={draw, rounded corners, minimum width=2.6cm, minimum height=1.2cm, align=center, font=\small},
  arrow/.style={-Stealth, thick},
  dashedarrow/.style={-Stealth, thick, dashed, color=red!70!black}
]

\node[box, fill=pink!30] (teacher) {Teacher\\Model};
\node[box, fill=yellow!30, right=of teacher, yshift=1.5cm] (block1) {Block 1};
\node[box, fill=yellow!30, right=of block1] (block2) {Block 2};
\node[box, fill=cyan!30, below=of block2, xshift=0cm] (student) {Student\\Model};

\draw[dashedarrow] (teacher.south east) to[bend left=10] node[midway, right, yshift=5pt, font=\small] {Logits} (student.north west);

\draw[arrow] (teacher.east) -- (block1.west);
\draw[arrow] (block1.east) -- (block2.west);
\draw[arrow] (block2.south) -- (student.north);

\node[font=\scriptsize, align=center] at ([xshift=0.3cm,yshift=-0.3cm]teacher.south) {Perception Calibration};
\node[font=\scriptsize, align=center] at ([yshift=0.7cm]block1.north) {Block-wise Logit\\Distillation};
\node[font=\scriptsize, align=center] at ([xshift=-0.3cm,yshift=-0.4cm]student.south) {Normalized\\Logits};

\end{tikzpicture}
\caption{Illustration of logit-based knowledge distillation with decoupled knowledge paths. Techniques shown include block-wise distillation, perception calibration, and temperature-based normalization.}
\label{fig:logit_kd}
\end{figure*}

Logit-based knowledge distillation (KD) has garnered significant attention in recent years and several studies have demonstrated its effectiveness in various scenarios. The potential of logit-based methods has been rediscovered, with \citep{2403.13512} and \citep{2411.01547} showing promising results, outperforming state-of-the-art feature-based methods in some cases. These findings highlight the importance of re-examining logit-based approaches that were initially overshadowed by feature-based methods. The concept of decoupling dark knowledge via block-wise logit distillation, introduced by \citep{2411.01547}, enables more efficient and flexible knowledge transfer. Similarly, scale-decoupled distillation \citep{2403.13512} reformulates the classical KD loss into two parts, allowing for better knowledge transfer between the teacher and student models. LumiNet \citep{2310.03669} introduced the concept of perception to calibrate logits based on the model's representation capability, addressing overconfidence issues in logit-based distillation. NormKD \citep{2308.00520} proposed normalized logits for KD by customizing the temperature for each sample according to its logit distribution.

A common theme among these studies is the importance of addressing the overconfidence issues in logit-based distillation. Overconfidence can lead to poor performance and various solutions have been proposed to mitigate this issue, including calibration \citep{2310.03669} and normalization \citep{2308.00520}. Decoupling and reformulation of knowledge are also popular approaches to improve logit-based distillation, allowing for more efficient and flexible knowledge transfer \citep{2403.13512, 2411.01547}. The use of intermediate stepping-stone models or decoupled knowledge can help bridge the gap between teacher and student models, as demonstrated by \citep{2203.08679}. Although logit-based methods have shown promise, there are contrasting viewpoints on their effectiveness compared to feature-based methods. Some studies argue that logit-based methods can be as effective or even superior in certain scenarios \citep{2403.13512, 2411.01547}, whereas others propose combining logits with features for better performance \citep{2310.03669}. The role of temperature in logit-based distillation is also debated, with some studies arguing that a single fixed value is insufficient and proposing customized temperatures for each sample \citep{2308.00520}.

From a technical perspective, block-wise logit distillation \citep{2411.01547} involves gradually replacing teacher blocks as intermediate stepping stone models to bridge the gap between students and teachers. Scale-decoupled distillation \citep{2403.13512} decouples knowledge into two parts, enabling more efficient and flexible knowledge transfer. Perception calibration \citep{2310.03669} calibrates logits based on the model's representation capability, whereas normalized logits \citep{2308.00520} customize the temperature for each sample according to its logit distribution. Logit-based methods may still be less effective than feature-based methods in certain scenarios, and further research is required to fully understand their strengths and limitations \citep{2211.14773}. The choice of temperature and formulation of KD loss can significantly impact the performance, requiring careful tuning and experimentation \citep{2402.12030}. In addition, the scalability of logit-based methods to more complex models and tasks remains an open question requiring further investigation \citep{2410.16215}. The lack of a unified framework for logit-based KD may hinder the development of more effective methods, highlighting the need for continued research in this area \citep{2409.12345}.

Significant progress has been made in logit-based KD in recent years, with various studies demonstrating its effectiveness under different scenarios. Although there are limitations and challenges that need to be addressed, the key themes of decoupling, reformulation, and addressing overconfidence issues have emerged as crucial components of logit-based methods. Further research is necessary to fully understand the strengths and limitations of logit-based approaches and develop more effective methods for knowledge distillation.

\section{Feature-based Knowledge Distillation}

Feature-based knowledge distillation (KD) is a variant of the KD technique that focuses on transferring knowledge through intermediate features rather than just output logits. This approach has gained significant attention in recent years, with several papers proposing novel methods to improve feature alignment between teacher and student models. For instance, \citep{2411.01547} proposed a block-wise logit distillation framework that combines the strengths of logit-based and feature-based methods, demonstrating comparable or superior results to those of state-of-the-art distillation methods. Similarly, \citep{2211.14773} revisited logit-based KD and proposed a class-aware logit knowledge distillation method that extended logit distillation to both the instance-level and class-level, outperforming several prevailing logit-based and feature-based methods. A common theme among these studies is the importance of improving the feature alignment between teacher and student models. A standard approach involves minimizing the $L_2$ distance \citep{dao2021knowledge} between intermediate features:

\begin{equation}
\mathcal{L}_{\text{feat}} = \frac{1}{N} \sum_{i=1}^{N} \left\| f_t^{(i)} - f_s^{(i)} \right\|_2^2,
\end{equation}
where $f_t^{(i)}$ and $f_s^{(i)}$ denote the feature representations from the teacher and student, respectively, for the $i$th training instance. \citep{2305.17007} propose a novel loss term, ND loss, that encourages the student model to produce large-norm features and aligns the direction of student features with teacher class-means, leading to state-of-the-art performance on several benchmarks. This highlights the significance of regularizing feature norms and directions in knowledge distillation. Furthermore, \citep{2102.02973} proposed an attention-based feature matching approach, demonstrating the effectiveness of using attention mechanisms to align features between teacher and student models. \citep{1912.10850} examined various KD strategies for simple classification tasks and found that many feature-based methods are difficult to reproduce and lack generalizability. This suggests a need for simpler, more effective, and reproducible distillation methods that can be applied to a wide range of tasks and datasets. Additionally, \citep{2112.10047} highlighted the importance of controlling the quality of distillation in response-based network compression, emphasizing the need to carefully consider the trade-offs between compression ratio, accuracy, and computational complexity.

The analysis of recent studies also reveals contrasting viewpoints and approaches to KD. For instance, \citep{2211.14773} argued that logit-based methods can be effective and even superior to feature-based methods in certain cases, whereas \citep{2411.01547} proposed a hybrid approach that combined the strengths of both logit-based and feature-based methods. This highlights the importance of considering specific characteristics of the task and dataset when selecting a distillation method. \citep{2308.00520} proposed a normalized logit approach for KD, demonstrating the effectiveness of using normalized logits to improve feature alignment.

In terms of technical details and methodologies, several papers have proposed novel architectures or loss functions to improve feature alignment. For example, \citep{2103.14337} proposed a hands-on guidance approach for distilling object detectors, whereas \citep{2403.05894} introduced a frequency attention mechanism for KD. These advancements have led to ongoing efforts to develop effective and efficient KD methods.

Overall, the studies analyzed in this section demonstrate the significance of feature alignment, norm, and direction in KD. They also highlight the need for simple, effective, and reproducible distillation methods that can be applied to a wide range of tasks and datasets. As research in this area continues to evolve, we are likely to see further developments in feature-based KD, including the proposal of novel architectures, loss functions, and methodologies that can improve the efficiency and effectiveness of these methods.

\section{Attention-based Knowledge Distillation}

Attention-based knowledge distillation has emerged as a significant area of research in the field of KD, and several studies have demonstrated its effectiveness in improving the performance of student models \citep{2304.04615, 2403.05894, 2305.15032, 2407.02775}. The use of attention mechanisms in KD has shown promising results, particularly for capturing the global information necessary for effective knowledge transfer \citep{2403.05894}. For example, frequency-attention-based KD can capture global information by adjusting the frequencies of student features under the guidance of teacher features \citep{2403.05894}. This approach has been found to be effective in improving model performance, especially when combined with other techniques such as multilevel knowledge distillation \citep{2407.02775}.

The application of attention mechanisms in KD can take various forms including attention and hidden state transfers \citep{2305.15032}. Attention-based feature matching is another effective approach for determining competent links between teacher and student features without manual selection, which leads to better model compression and transfer learning results \citep{2102.02973}. Multi-level KD, which explores relation-level knowledge and allows for flexible student attention head number setting, has also been found to improve model performance \citep{2407.02775}. Emphasis on capturing global information or relation-level knowledge for effective knowledge transfer is a common theme among these studies, highlighting the importance of considering the relationships between different features and levels of representation.

\begin{figure}[ht]
\centering
\begin{tikzpicture}
\small
\begin{axis}[
    ybar,
    bar width=8pt,
    width=8cm,
    height=7cm,
    enlarge x limits=0.15,
    ylabel={Metric Value},
    symbolic x coords={Top-1 Acc, Top-5 Acc, FLOPs, Params, Transfer Score, Calib. Error},
    xtick=data,
    xtick align=inside,
    xticklabel style={rotate=30, anchor=east, yshift=-6pt},
    ymin=0,
    legend style={at={(0.5,-0.25)}, anchor=north, legend columns=-1},
    nodes near coords,
    every node near coord/.append style={rotate=30, font=\scriptsize},
    grid=major,
    grid style={dashed,gray!30},
    area legend,
]
\addplot+[fill=blue!40, bar shift=-6pt] coordinates {
    (Top-1 Acc,76.2)
    (Top-5 Acc,93.1)
    (FLOPs,1.9)
    (Params,11.2)
    (Transfer Score,0.72)
    (Calib. Error,0.14)
};

\addplot+[fill=orange!60, bar shift=6pt] coordinates {
    (Top-1 Acc,77.8)
    (Top-5 Acc,94.0)
    (FLOPs,2.3)
    (Params,12.5)
    (Transfer Score,0.78)
    (Calib. Error,0.11)
};
\footnotesize
\legend{Logit-based KD, Attention-based KD}
\end{axis}
\end{tikzpicture}
\caption{Comparison of Logit-based and Attention-based Knowledge Distillation across multiple evaluation metrics.}
\label{fig:kd_comparison}
\end{figure}

The technical details and methodologies employed in attention-based KD have varied across studies. For example, the frequency attention module used in \citep{2403.05894} is a learnable global filter that adjusts the frequencies of student features under the guidance of teacher features. In contrast, the attention-based meta-network proposed by \citep{2102.02973} learns the relative similarities between features and applies the identified similarities to control the distillation intensities. The choice of distillation objectives, such as attention transfer, hidden state transfer, and vanilla KD, also varies across studies \citep{2305.15032}. Despite these differences, the use of attention mechanisms in KD has consistently improved model performance, particularly in task-specific settings.

Large-scale experiments are often required to evaluate the effectiveness of these methods \citep{2304.04615, 2403.05894, 2305.15032, 2407.02775}. In addition, determining the optimal attention mechanism or objective for specific tasks or datasets is challenging \citep{2305.15032}. The potential computational cost of attention mechanisms in KD is another concern, particularly for large-scale models \citep{2102.02973, 2403.05894}. Despite these challenges and as shown Figure \ref{fig:kd_comparison}, it is evident that attention-based knowledge distillation consistently outperforms the logit-based counterpart across most metrics, demonstrating superior accuracy, better calibration, and stronger transferability. The development of attention-based KD methods has significant implications in the field of deep learning, enabling a more efficient and effective transfer of knowledge between models. 

In conclusion, attention-based KD is a rapidly evolving area of research with significant potential to improve the performance of student models. The use of attention mechanisms in KD has been consistently shown to capture the global information necessary for effective knowledge transfer, particularly when combined with other techniques, such as multi-level KD. Despite the limitations and challenges associated with these methods, the development of attention-based KD has significant implications for the field of deep learning, enabling more efficient and effective transfer of knowledge between models. Future research should focus on addressing the challenges associated with attention-based KD, such as determining the optimal attention mechanism or objective for specific tasks or datasets, and exploring new applications and extensions of these methods \citep{2102.02973, 1912.10850, 2305.10893}.

\input{forest}

\section{Image Classification Applications}

Image classification with knowledge distillation (KD) has been a vibrant area of research, and numerous studies have explored its potential for improving the performance of deep neural networks. A key finding in this domain is the effectiveness of using unlabeled mismatched images for KD, as demonstrated by \citep{1703.07131}, which highlights the importance of stimulus complexity in this process. Furthermore, \citep{2303.08360} introduced a novel distillation method based on Class Activation Maps (CAMs) for multi-label classification, outperforming traditional logit-based and feature-based methods. This underscores the challenges associated with applying KD to multi-label classification tasks and the need for innovative approaches.

Recent surveys such as \citep{2304.04615,2404.00936} have compiled various KD strategies, including teaching-assistant distillation, curriculum distillation, mask distillation, and decoupling distillation, showing promising results in improving KD. The study in \citep{2209.15555} provided a unified framework for affinity-based KD, modularizing the process into three components: affinity, normalization, and loss, and evaluating numerous distillation objectives for image classification. This emphasis on unified frameworks was echoed by \citep{1912.10850}, who surveyed various KD strategies for simple classification tasks and highlighted the difficulty of reproducing state-of-the-art accuracy results.

The importance of stimulus complexity, as noted by \citep{1703.07131}, intersects with the exploration of different KD variants, including those discussed in \citep{2304.04615}. The challenges in multi-label classification, identified in \citep{2303.08360}, also reflect a broader theme within the field: the need for standardized evaluation protocols and unified frameworks to facilitate comparison and advancement. This is reinforced by both \citep{2209.15555} and \citep{1912.10850}, who advocate for more systematic approaches to KD.

Technical details, such as the use of CAMs by \citep{2303.08360} and affinity-based KD by \citep{2209.15555}, highlight the diversity of the methodologies being explored. The teacher-student learning frameworks discussed by \citep{2304.04615} further underscore the complexity and richness of the current research in this area. However, limitations, such as the reproducibility of state-of-the-art results, as highlighted in \citep{1912.10850}, and the generalizability of distillation techniques across different architectures and training settings, as noted in \citep{1912.10850} and \citep{2303.08360}, pose significant challenges.

The implications of these developments are multifaceted. On one hand, they suggest a vibrant field with considerable potential for innovation and improvement. However, they underscore the need for more rigorous evaluation standards and a deeper understanding of how different KD methods interact with various deep learning architectures and tasks. The connections between KD and other areas of deep learning, such as the use of synthetic data generated by diffusion models for KD, as explored by \citep{2305.12954}, further broadens the scope of potential applications and challenges.

Image classification with KD represents a dynamic and evolving field characterized by significant advancements and ongoing challenges. As research continues to explore new methodologies and address existing limitations, the potential of KD to enhance deep learning performance in image classification and beyond remains considerable. Future studies will likely focus on developing more generalized and robust distillation techniques as well as exploring the application of KD in a wider range of tasks and domains, potentially leveraging insights from \citep{2110.04100,2105.02788} and other related works to drive innovation forward.

\section{Object Detection and Segmentation}

Object detection and segmentation are core tasks in computer vision, and KD has been proven to be effective in enhancing accuracy and efficiency. Recent studies have investigated various KD strategies for these tasks, resulting in notable improvements. The study in \citep{2406.06999} introduced a novel feature-based distillation method that leverages knowledge uncertainty and achieves state-of-the-art performance on the COCO dataset.

A recurring focus in recent research is the application of feature-based distillation methods for object detection. For instance, the investigations in \citep{2404.01699, 2308.09105} illustrate the effectiveness of transferring knowledge from teacher models to student models through feature-based approaches. In addition, \citep{2110.09674} presented a novel KD framework that narrowed the feature gap between the teacher and student models, which is particularly relevant for UAV-based object detection.

Another key development involves mitigating the uncertainty and noise in the teacher model's knowledge. For instance, the approach in \citep{2203.05469} concentrates on the teacher model's most predictive regions, achieving performance gains over the existing KD baselines. Similarly, \citep{2110.09674} highlighted the importance of addressing uncertainty and noise in teachers' knowledge to achieve a more effective distillation.

Balancing classification and regression tasks were common themes in these studies. Studies by \citep{2102.12252, 2112.04840} demonstrated the importance of considering both classification and regression tasks in object detection, and \citep{2108.07482} introduced a general distillation framework designed to maintain the equilibrium between classification and regression tasks.

Various architectures and distillation methods have been employed to produce different models. For instance, \citep{2201.11097} adopted a ResNet-50 backbone for object detection in autonomous driving scenarios, whereas \citep{2308.09105} explored a multi-scale feature learning approach for person re-identification. In addition, \citep{2406.06999} examined specialized loss functions to optimize the performance.

Computational resources are a cause for concern because KD can be computationally expensive \citep{2308.09105}. In addition, alleviating the feature gap between the teacher and student models remains a challenge, particularly in domains such as UAV-based object detection \citep{2408.11407}. Uncertainty estimation and noise robustness are also essential considerations, because they can significantly affect the performance of the student model \citep{2203.05469}.

In conclusion, KD has emerged as a powerful approach for improving object detection and segmentation tasks. Recent studies have highlighted key themes, such as feature-based distillation methods, addressing uncertainty and noise, and balancing classification and regression tasks. Although challenges and limitations remain, advancements in this area have significant implications for computer vision applications, including autonomous driving, surveillance, and robotics. As research continues to evolve, KD is likely to play an increasingly important role in improving the accuracy and efficiency of object detection and segmentation systems.

\section{Language Modeling}

Language modeling with knowledge distillation (KD) has emerged as a crucial technique for compressing large language models while maintaining their accuracy, as demonstrated in various studies \citep{2407.01885,2404.14827,2407.02775}. The core idea behind KD is to transfer knowledge from a pretrained teacher model to a smaller student model, thereby reducing computational requirements and improving inference speed without significantly compromising performance. A comprehensive survey of KD for large language models \citep{2407.01885} provides an overview of the techniques, evaluation tasks, and applications, highlighting the effectiveness of this approach.

Recent studies have explored different distillation methods, including sentence-level, token-level, and hybrid approaches. For instance, \citep{2404.14827} introduced a novel hybrid method that combines token-level and sentence-level distillation, outperforming the individual methods and previous studies. \citep{2407.02775} proposed a multi-level KD method that improves model performance and allows for flexible student attention head number settings. Furthermore, \citep{2109.08359} presented a new KD objective that transfers contextual knowledge via word relations and layer transforming relations without restrictions on architectural changes between teacher and student models.

The importance of selecting an appropriate distillation method depending on scenario complexity has also been emphasized in the literature. For example, \citep{2404.14827} argued that token-level distillation is better suited to simple scenarios, whereas sentence-level distillation excels in complex scenarios. However, other studies have not explicitly made this distinction, highlighting the need for further research on task-agnostic distillation and its applications \citep{2302.09632}. The use of pretrained language models as teachers and the exploration of various student models, including those with smaller sizes or different architectures, are also a common theme in the literature \citep{2407.01885,2211.01071}.

Technical details and methodologies vary across studies, with KD techniques including sentence-level, token-level, hybrid, multi-level, and task-agnostic distillation. Model architectures such as BERT, pretrained language models, and various student models with smaller sizes or different architectures have also been employed \citep{2301.08130,2407.02775}. Benchmarks, such as the GLUE benchmark, extractive question answering, and other language understanding tasks, are used to assess the performance of distilled models \citep{2404.14827,2112.02505}.

The complexity of selecting the appropriate distillation method depending on the scenario, maintaining a small discrepancy between the teacher's and student's predictions during the distillation process, and defining the complexity level of a given scenario are challenges faced by researchers \citep{2406.16524,2302.00444}. Furthermore, the current KD approaches have limitations in terms of scalability, flexibility, and adaptability to different models and tasks, thereby highlighting the need for further research in this area \citep{2410.16215,2307.15190}.

However, language modeling with KD has become a vital technique for compressing large language models while maintaining their accuracy. The literature highlights the importance of selecting an appropriate distillation method, the use of pre-trained language models as teachers, and the exploration of various student models. However, limitations and challenges remain, emphasizing the need for further research on task-agnostic distillation, scalability, flexibility, and adaptability to various models and tasks. As the field continues to evolve, the KD is likely to play an increasingly important role in the development of efficient and effective language models \citep{2407.01885,2302.09632}.

\section{Text Classification and Sentiment Analysis}

Text classification and sentiment analysis are pivotal tasks in natural language processing (NLP), and KD has emerged as a significant technique for enhancing model efficiency and accuracy. Recent studies have introduced variants of KD, such as teaching-assistant distillation, curriculum distillation, mask distillation, and decoupling distillation, to improve performance by modifying the learning process or adding new components \citep{2304.04615}. Notably, the development of Knowledge Distillation Semi-supervised Topic Modeling (KDSTM) offers an efficient method for text classification that requires no pre-trained embeddings and few labeled documents, making it particularly suitable for resource-constrained settings \citep{2307.01878}.

The effectiveness of KD in leveraging unlabeled data has also been demonstrated, with research showing that stimulus complexity is crucial for good performance \citep{1703.07131}. Furthermore, there is an ongoing debate regarding the choice between sentence- and token-level distillation, with studies suggesting that token-level distillation is better suited for simple scenarios, whereas sentence-level distillation excels in more complex scenarios \citep{2404.14827}. This has led to proposals for hybrid methods that combine both approaches through innovative mechanisms such as gating.

Experimental validation has been stressed as a critical aspect of KD research, with many studies highlighting the need for thorough and consistent evaluation methodologies to substantiate the claims of improved performance \citep{1912.10850}. However, reproducibility issues have been identified as a significant challenge, particularly for methods that employ feature distillation, underscoring the importance of standardized model architectures and training schedules \citep{2011.14554}.

The use of deep neural networks (DNNs) and pre-trained language models such as BERT has been prevalent in KD for text classification and sentiment analysis \citep{2002.12620}. Moreover, data augmentation techniques have been recognized as crucial for enhancing the distillation outcomes \citep{2110.04741}. The proposal of adaptive distillation methods that can aggregate knowledge from multiple paths represents an innovative approach to efficient distillation \citep{2110.09674}.

Reliance on large amounts of labeled data for traditional KD methods remains a challenge. Additionally, it is difficult to define the scenario complexity of different distillation approaches \citep{1812.11587}. The relationship between KD and label smoothing has also been debated, with some studies suggesting that they may be more closely related than initially thought \citep{2301.12609}.

The core of research on text classification and sentiment analysis with KD contributes significantly to the field by introducing new methodologies, emphasizing the importance of experimental validation, and discussing challenges, such as reproducibility and scenario complexity. These findings are invaluable for advancing the state of KD for more efficient and accurate model training across various tasks and datasets \citep{2407.01885}. As the field continues to evolve, addressing current limitations and exploring new avenues, such as hybrid distillation methods and adaptive techniques, will be crucial for realizing the full potential of KD in NLP applications \citep{2407.13952}.

\input{comp_table}

\section{Recent Progress and Challenges}

Knowledge distillation (KD) has undergone significant advancements in recent years, with a plethora of research papers contributing to its growth, as shown in Figure \ref{fig:forest}. As evident from recent studies \citep{1912.13179,2304.04615,2011.14554,2109.12507,2410.14741}, KD has become a vital component in the development of deep neural networks, enabling the transfer of knowledge from complex teacher models to simpler student models. The introduction of new variants of KD, such as teaching-assistant distillation, curriculum distillation, mask distillation, and decoupling distillation \citep{2109.12507,2410.14741}, has improved KD performance by introducing additional components or changing the learning process.

Furthermore, research papers have proposed novel perspectives on knowledge quantity and its influence on distillation procedures, leading to the development of the partial-to-whole knowledge distillation (PWKD) paradigm \citep{2109.12507}. The decoupling of the Kullback-Leibler divergence into three distinct components, Binary Classification Divergence (BCD), Strong Correlation Divergence (SCD), and Weak Correlation Divergence (WCD) \citep{2410.14741}, has further facilitated the development of a Correlation-Aware Knowledge Distillation (CAKD) framework. These advancements have been complemented by the development of general models for KD that summarize various techniques and methods in the field \citep{1912.13179,2011.14554}.

A common theme among these studies was the emphasis on improving KD performance, with most focusing on introducing new techniques, components, or perspectives to enhance the learning process \citep{2304.04615,2109.12507}. The importance of considering both the knowledge quantity and quality in the distillation procedure has also been highlighted, with studies emphasizing the need for a deeper understanding of the distillation process \citep{1912.13179,2410.14741}. Additionally, the decomposition and analysis of existing concepts, such as the Kullback-Leibler divergence, have enabled researchers to gain a deeper understanding of the distillation process \citep{2410.14741}.

Moreover, the analyzed studies offer divergent perspectives on KD, encompassing variations in methodology and diverse application domains \citep{1912.13179,2304.04615,2109.12507}. While concentrating on improving the distillation performance, others have proposed new frameworks or models that extend existing techniques \citep{2011.14554,2410.14741}. The technical approaches in these studies are equally diverse, such as the decoupling of the Kullback-Leibler divergence to introduce additional components, such as the CAKD framework \citep{2410.14741}.

The limitations and challenges of some KDs remain, including limited understanding of the KD process and its components \citep{1912.13179,2011.14554}. The need for more efficient and scalable KD methods, particularly for large-scale applications, has also been highlighted \citep{2304.04615,2410.14741}. Furthermore, the importance of explaining and analyzing the performance gain achieved through KD cannot be overstated, as it is essential to understand the underlying mechanisms of knowledge transfer \citep{1912.13179,2011.14554}.

The current state of KD research is characterized by a vibrant and dynamic community, with researchers continually exploring new techniques, components, and perspectives to enhance the learning process. As the field continues to evolve, it is essential to address the limitations and challenges that remain, including the need for a deeper understanding of the distillation process, more efficient and scalable methods, and explainability and analysis of the performance gains. Thus, the KD can continue to play a vital role in the development of deep neural networks, enabling the creation of more efficient, effective, and intelligent systems \citep{1912.13179,2304.04615,2011.14554,2109.12507,2410.14741}. As research in this area continues to advance, we expect to observe significant improvements in the performance and efficiency of KD, leading to new applications and innovations in the field of deep learning \citep{1912.13179,2011.14554,2410.14741}. Ultimately, KD holds great promise for potential applications in areas such as computer vision, natural language processing, and robotics \citep{2304.04615,2109.12507}.

\section{Advancements and Emerging Trends}

Note that the above outline can be modified based on the specific requirements or focus areas of the survey paper, as the field of KD is rapidly evolving, with new variants and techniques being proposed (as shown in Table \ref{tab:kd_comparison_sorted}). For instance, recent studies have explored teaching-assistant distillation \citep{2304.04615}, curriculum distillation, mask distillation, and decoupling distillation, which have shown promising results in improving KD performance. Moreover, the concepts of knowledge decomposition and the partial-to-whole knowledge distillation (PWKD) paradigm highlight the significance of knowledge quantity in the distillation process \citep{2109.12507}. The importance of Correlation-Aware Knowledge Distillation (CAKD) has also been emphasized, in which the CAKD framework prioritizes the most influential facets of distillation components, leading to improved knowledge transfer \citep{2410.14741}. Additionally, spot-adaptive knowledge distillation (SAKD) adaptively determines the distillation spots in the teacher network per sample at every training iteration to improve distillation performance \citep{2205.02399}. These developments underscore the need for a comprehensive survey that can accommodate a diverse range of techniques and applications in KD, including improved KD for pretrained language models via knowledge selection \citep{2302.00444}, comparable KD in semantic image segmentation \citep{2309.03659}, and control of the quality of distillation in response-based network compression \citep{2112.10047}. The survey should also consider the technical details and methodologies employed in these studies, such as neural network architectures, distillation losses, and training procedures, to provide a thorough understanding of the current state of KD research. Furthermore, the limitations and challenges faced in this field, including reproducibility \citep{1912.10850}, scalability \citep{2205.10003}, and interpretability, should be discussed to identify areas for future research and development. By providing a structured overview of the key findings, common themes, contrasting viewpoints, technical details, and limitations in the field of KD, this survey aimed to contribute to ongoing efforts to advance our understanding of this important area in machine learning.

\section{Conclusion}

Knowledge distillation has advanced considerably in recent years, with novel strategies, such as teaching assistant distillation, curriculum distillation, mask distillation, decoupling distillation, correlation-aware knowledge distillation, and partial knowledge distillation proposed to enhance teacher-student learning. These approaches either introduce new components or reformulate the learning process, underscoring the need for precise control over distillation quality. Notably, CAKD framework decouples the Kullback-Leibler divergence into BCD, SCD, and WCD, while PWKD restructures the teacher into weight-sharing subnetworks. Several studies also emphasise the role of teaching assistants, similarity-based guidance, and the integration of classical distillation with augmentation schemes. However, challenges remain chiefly reproducibility, generalisability, and standardised evaluation, especially in feature-based distillation settings. As research continues to explore new architectures and training configurations, the field must prioritise robust, transferable techniques to ensure progress across applications in computer vision and natural language processing.

\bibliographystyle{apalike}
\bibliography{cite}

\end{document}

%% file: kd.tex
\begin{figure*}[t]
\centering
\begin{tikzpicture}[
    node distance=2.5cm and 3cm,
    every node/.style={align=center, font=\sffamily},
    box/.style={draw, thick, rounded corners=5pt, minimum width=2.2cm, minimum height=1.3cm},
    teacher/.style={box, fill=blue!15, text=blue!80!black},
    student/.style={box, fill=green!15, text=green!60!black},
    data/.style={box, fill=gray!10, text=gray!80!black},
    output/.style={box, fill=orange!15, text=orange!70!black},
    arrow/.style={thick, ->, >=stealth},
    distill/.style={thick, ->, >=stealth, draw=red!60}
]

\node[teacher] (teacher) {Teacher Model};
\node[student, below=3cm of teacher] (student) {Student Model};
\node[data, left=of teacher] (data) {Input Data};
\node[output, right=of teacher] (output) {Hard Labels};

\draw[arrow, draw=blue!60] (data.east) -- (teacher.west);
\draw[arrow, draw=blue!60] (teacher.east) -- (output.west);
\draw[arrow, draw=green!60] (data.south east) to[out=0, in=180] (student.west);

\draw[distill] (teacher) -- node[midway, left, text=red!60] {Knowledge\\Distillation} (student);

\node[below right=1cm and 0.2cm of teacher, text=red!60, draw=red!60, 
      dashed, rounded corners, fill=red!5, minimum width=3.5cm, minimum height=1.5cm] (soft) {Soft Labels\\$p_i = \frac{\exp(z_i/T)}{\sum_j \exp(z_j/T)}$};
\draw[distill, dashed] (teacher.east) to[out=0, in=90] (soft.north);
\draw[distill, dashed] (soft.south) to[out=-90, in=0] (student.east);

\node[above=0.1cm of teacher, font=\scriptsize\itshape, text=blue!60!black] {Large, Complex, High Performance};
\node[below=0.1cm of student, font=\scriptsize\itshape, text=green!60!black] {Small, Efficient, Fast Inference};

\end{tikzpicture}
\caption{An overview of the Knowledge Distillation (KD) framework, where a large, high-performance teacher model transfers knowledge to a lightweight student model. Soft labels derived from the teacher’s softened logits guide the student’s learning alongside hard labels.}
\label{fig:distillation}
\end{figure*}

%% file: timeline.tex
\begin{figure*}[t]
\centering
\begin{tikzpicture}[
  timeline/.style={draw=gray!60, thick},
  event/.style={circle, draw=black, fill=blue!20, minimum size=5mm},
  label/.style={font=\footnotesize, align=center}
]
\draw[timeline] (0,0) -- (16,0);

\foreach \x/\year/\model [count=\i] in {
  0/2006/Model Compression (Bucilă et al.),
  2/2015/Knowledge Distillation (Hinton et al.),
  4/2018/Teaching Assistant KD (TAKD),
  6/2020/TextBrewer for NLP Tasks,
  8/2021/Correlation-Aware KD (CAKD),
  10/2022/NormKD \& Class-aware Logit KD,
  12/2023/Block-wise Logit Distillation,
  14/2024/Scale Decoupled Distillation,
  16/2025/LLM-driven KD for Dynamic Graphs
} {
  \node[event] at (\x,0) {};
  
  \pgfmathparse{mod(\i,2)==0}
  \ifnum\pgfmathresult=1
    \node[label, below=0.3cm] at (\x,0) {\year};
    \node[label, above=0.3cm, text width=2.8cm] at (\x,0) {\model};
  \else
    \node[label, below=0.3cm] at (\x,0) {\year};
    \node[label, below=0.7cm, text width=2.8cm] at (\x,0) {\model};
  \fi
}
\end{tikzpicture}
\caption{Timeline of major developments in Knowledge Distillation (KD).}
\label{fig:kd_timeline}
\end{figure*}

%% file: forest.tex
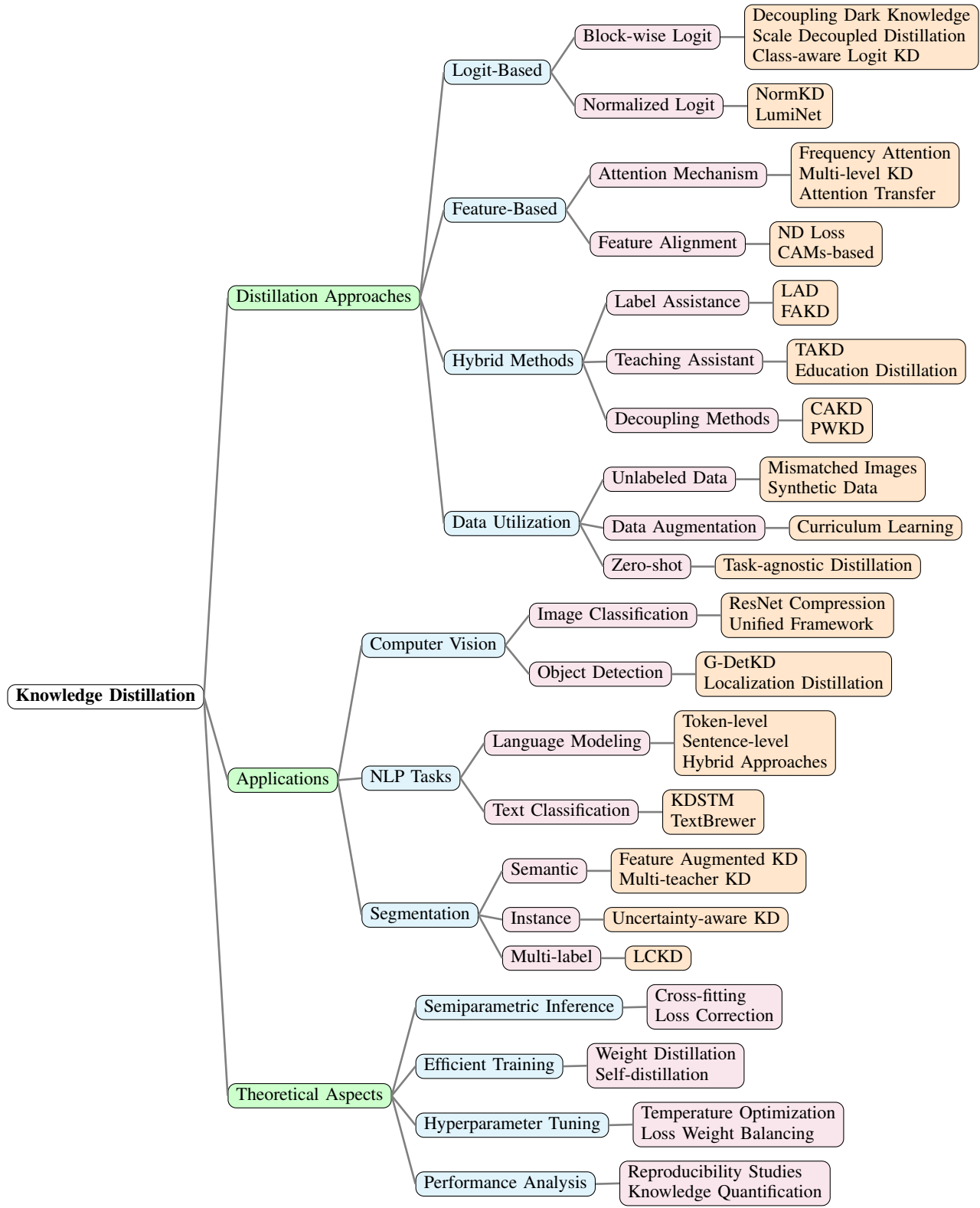
\begin{figure*}
  \centering
\begin{forest}
  for tree={
    grow=east,
    reversed=true,
    anchor=base west,
    parent anchor=east,
    child anchor=west,
    base=left,
    rectangle,
    draw=black,
    rounded corners,
    align=left,
    minimum width=3em,
    edge+={gray, line width=1pt},
    inner xsep=4pt,
    inner ysep=1pt,
    font=\small,
  },
  where level=1{fill=green!20}{},
  where level=2{fill=cyan!10}{},
  where level=3{yshift=0.3pt, fill=purple!10}{},
  where level=4{yshift=0.3pt, fill=orange!20}{},
  where level=5{yshift=0.3pt, fill=gray!10}{},
  [\textbf{Knowledge Distillation},
    [Distillation Approaches,
        [Logit-Based,
            [Block-wise Logit,
                [Decoupling Dark Knowledge \\
                 Scale Decoupled Distillation \\
                 Class-aware Logit KD
                ]
            ]
            [Normalized Logit,
                [NormKD \\
                 LumiNet
                ]
            ]
        ]
        [Feature-Based,
            [Attention Mechanism,
                [Frequency Attention \\
                 Multi-level KD \\
                 Attention Transfer
                ]
            ]
            [Feature Alignment,
                [ND Loss \\
                 CAMs-based
                ]
            ]
        ]
        [Hybrid Methods,
            [Label Assistance,
                [LAD \\
                 FAKD
                ]
            ]
            [Teaching Assistant,
                [TAKD \\
                 Education Distillation
                ]
            ]
            [Decoupling Methods,
                [CAKD \\
                 PWKD
                ]
            ]
        ]
        [Data Utilization,
            [Unlabeled Data,
                [Mismatched Images \\
                 Synthetic Data
                ]
            ]
            [Data Augmentation,
                [Curriculum Learning
                ]
            ]
            [Zero-shot,
                [Task-agnostic Distillation
                ]
            ]
        ]
    ]
    [Applications,
        [Computer Vision,
            [Image Classification,
                [ResNet Compression \\
                 Unified Framework
                ]
            ]
            [Object Detection,
                [G-DetKD \\
                 Localization Distillation
                ]
            ]
        ]
        [NLP Tasks,
            [Language Modeling,
                [Token-level \\
                 Sentence-level \\
                 Hybrid Approaches
                ]
            ]
            [Text Classification,
                [KDSTM \\
                 TextBrewer
                ]
            ]
        ]
        [Segmentation,
            [Semantic,
               [Feature Augmented KD \\
                Multi-teacher KD
               ]
            ]
            [Instance,
              [Uncertainty-aware KD
              ]
            ]
            [Multi-label,
              [LCKD
              ]
            ]
        ]
    ]
    [Theoretical Aspects,
        [Semiparametric Inference,
            [Cross-fitting \\
             Loss Correction
            ]
        ]
        [Efficient Training,
            [Weight Distillation \\
             Self-distillation
            ]
        ]
        [Hyperparameter Tuning,
            [Temperature Optimization \\
             Loss Weight Balancing
            ]
        ]
        [Performance Analysis,
            [Reproducibility Studies \\
             Knowledge Quantification
            ]
        ]
    ]
  ]
\end{forest}
\caption{\textbf{Taxonomy of Knowledge Distillation (KD) for different artificial intelligence tasks.}}
\label{fig:forest}
\end{figure*}

%% file: comp_table.tex
\begin{table*}[t]
\caption{Comparison of Knowledge Distillation Methods (Sorted by Year)}
\label{tab:kd_comparison_sorted}
\centering
\resizebox{\textwidth}{!}{%
\begin{tabular}{@{}lllllp{3.5cm}p{3.5cm}@{}}
\toprule
\textbf{Method} & \textbf{Reference} & \textbf{Year} & \textbf{Main Application} & \textbf{Core Technology} & \textbf{Advantages} & \textbf{Limitations} \\
\midrule
Classic Knowledge Distillation & Hinton et al. & 2015 & Image Classification & Soft target knowledge transfer & Simple implementation, widely applicable & Performance limited by teacher model \\
Teaching Assistant Distillation & Mirzadeh et al. & 2020 & Various tasks & Intermediate model bridging & Effectively bridges teacher-student complexity gap & Requires additional model overhead \\
Curriculum Distillation & Gao et al. & 2020 & Image Classification & Progressive curriculum-based learning & Improves knowledge transfer efficiency & Complex curriculum design \\
Mask Distillation & Shi et al. & 2020 & Computer Vision & Selective feature transfer & Precise knowledge extraction & High mask design requirements \\
Weight Distillation & Yim et al. & 2020 & Various tasks & Parameter-level knowledge transfer & Significantly reduced training time (1.88-2.94×) & High implementation complexity \\
Zero-shot Knowledge Distillation & Micaelli et al. & 2020 & NLP & Learning without task-specific data & Solves data privacy issues & May experience performance degradation \\
Partial to Whole Knowledge Distillation (PWKD) & Kim et al. & 2021 & Image Classification & Weight-sharing subnetworks & Improves knowledge decomposition and transfer & High computational resource requirements \\
Self-Distillation & Xu et al. & 2021 & Various tasks & Self-optimizing iteration & No pre-trained teacher model needed & Performance ceiling may be limited \\
Decoupled Distillation & Zhao et al. & 2022 & Various tasks & Decoupling logits and features & High flexibility and adaptability & High implementation complexity \\
Class-aware Logit Knowledge Distillation (CLKD) & Zhou et al. & 2022 & Classification tasks & Instance-level and class-level combination & Enhanced semantic information extraction capability & Requires additional supervisory signals \\
Correlation-Aware Knowledge Distillation (CAKD) & Bi et al. & 2023 & Classification tasks & KL divergence three-element decoupling & Prioritizes high-impact distillation components & High computational complexity \\
Block-wise Logit Distillation & Wang et al. & 2023 & Various tasks & Block replacement feature alignment & Establishes effective intermediate transition models & High design complexity \\
Education Distillation & Hao et al. & 2023 & Various tasks & School-style incremental learning & Good progressive knowledge acquisition & Complex training process design \\
Frequency Attention Distillation & Wei et al. & 2024 & Computer Vision & Frequency domain feature adjustment & Strong global information capture ability & Large computational overhead \\
Sentence-level Distillation & Li et al. & 2024 & NLP & Sentence-level knowledge transfer & Excellent performance in complex NLP scenarios & May be suboptimal for simple scenarios \\
Token-level Distillation & Li et al. & 2024 & NLP & Token-level knowledge transfer & Excellent performance in simple NLP scenarios & May be suboptimal for complex scenarios \\
Hybrid Distillation & Li et al. & 2024 & NLP & Sentence-level + token-level combination & Strong scenario adaptability and universality & High implementation complexity \\
Feature-based Distillation & Yi et al. & 2024 & Computer Vision/NLP & Intermediate feature alignment & Captures rich representation information & Difficult to reproduce, unstable performance \\
Logit-based Distillation & Yu et al. & 2024 & Various tasks & Output logit alignment & Simple implementation, stable training & May not transfer information sufficiently \\
\bottomrule
\end{tabular}
}
\end{table*}